\documentclass[10pt,twocolumn,letterpaper]{article}

\usepackage{cvpr}              

\usepackage[dvipsnames]{xcolor}

\definecolor{cvprblue}{rgb}{0.21,0.49,0.74}
\usepackage[pagebackref,breaklinks,colorlinks,citecolor=cvprblue]{hyperref}

\def\paperID{376} 
\def\confName{3DV\xspace}
\def\confYear{2024\xspace}

\usepackage[utf8]{inputenc} 
\usepackage[T1]{fontenc}    
\usepackage{hyperref}       
\usepackage{url}            
\usepackage{booktabs}       
\usepackage{amsfonts}       
\usepackage{nicefrac}       
\usepackage{microtype}      
\usepackage{xcolor}         
\usepackage{array, multirow}
\usepackage{rotating}

\usepackage{times,epsfig,graphicx}
\usepackage{algorithm2e,algpseudocode}
\usepackage{verbatim}
\usepackage{tabularx}
\usepackage{xspace}
\usepackage{bbm}
\usepackage{bm}
\usepackage{booktabs}  

\usepackage{amssymb,amsmath}
\usepackage{nicematrix}

\newlength{\smallimage}
        \definecolor{rel}{rgb}{.1,.6,.2}
        \definecolor{nrl}{rgb}{1,1,1}
        \definecolor{qim}{rgb}{1,1,1}

\makeatletter

\def\eg{\emph{e.g.\,}}

\def\be{\begin{equation}}
\def\ee{\end{equation}}
\def\bea{\begin{eqnarray}}
\def\eea{\end{eqnarray}}
\def\ben{\begin{eqnarray*}}
\def\een{\end{eqnarray*}}

\def\bi{\begin{itemize}}
\def\ei{\end{itemize}}

\newcommand{\btab}[1]{\begin{tabular}{#1}}
\newcommand{\etab}{\end{tabular}}
\newcommand{\ba}[1]{\begin{array}{#1}}
\newcommand{\ea}{\end{array}}

\DeclareMathOperator*{\argmax}{\mathrm{argmax}}

\def\<{\langle}
\def\>{\rangle}

\newcommand{\R}{\mathbb{R}}



\definecolor{DarkCoral}{rgb}{0.8, 0.36, 0.27}

\newcommand{\PAR}[1]{\vskip4pt \noindent{\bf #1.}}

\title{Self-Supervised Learning of Neural Implicit Feature Fields for Camera Pose Refinement}

\author{Maxime Pietrantoni$^{1,2}$
\and
Gabriela Csurka$^3$ \\
\and
Martin Humenberger$^3$
\and
Torsten Sattler$^2$
\and
$^1$ Faculty of Electrical Engineering, Czech Technical University in Prague \\
$^2$ Czech Institute of Informatics, Robotics and Cybernetics, Czech Technical University in Prague \\
$^3$ NAVER LABS Europe \\
{\tt\small \{firstname.lastname\}@cvut.cz},\tt\small \{firstname.lastname\}@naverlabs.com
}

\begin{document}
\maketitle

\begin{abstract}
Visual localization techniques rely upon some underlying scene representation to localize against. These representations can be explicit such as 3D SFM map or implicit, such as a neural network that learns to encode the scene. The former requires sparse feature extractors and matchers to build the scene representation. The latter might lack geometric grounding not capturing the 3D structure of the scene well enough. This paper proposes to jointly learn the scene representation along with a 3D dense feature field and a 2D feature extractor whose outputs are embedded in the same metric space. Through a contrastive framework we align this volumetric field with the image-based extractor and regularize the latter with a ranking loss from learned surface information. We learn the underlying geometry of the scene with an implicit field through volumetric rendering and design our feature field to leverage intermediate geometric information encoded in the implicit field. The resulting features are discriminative and robust to viewpoint change while maintaining rich encoded information. Visual localization is then achieved by aligning the image-based features and the rendered volumetric features. We show the effectiveness of our approach on real-world scenes, demonstrating that our approach outperforms prior and concurrent work on leveraging implicit scene representations for  localization.
\end{abstract}

\section{Introduction}

\begin{figure}[t]
\centering
\includegraphics[width=1.\linewidth]{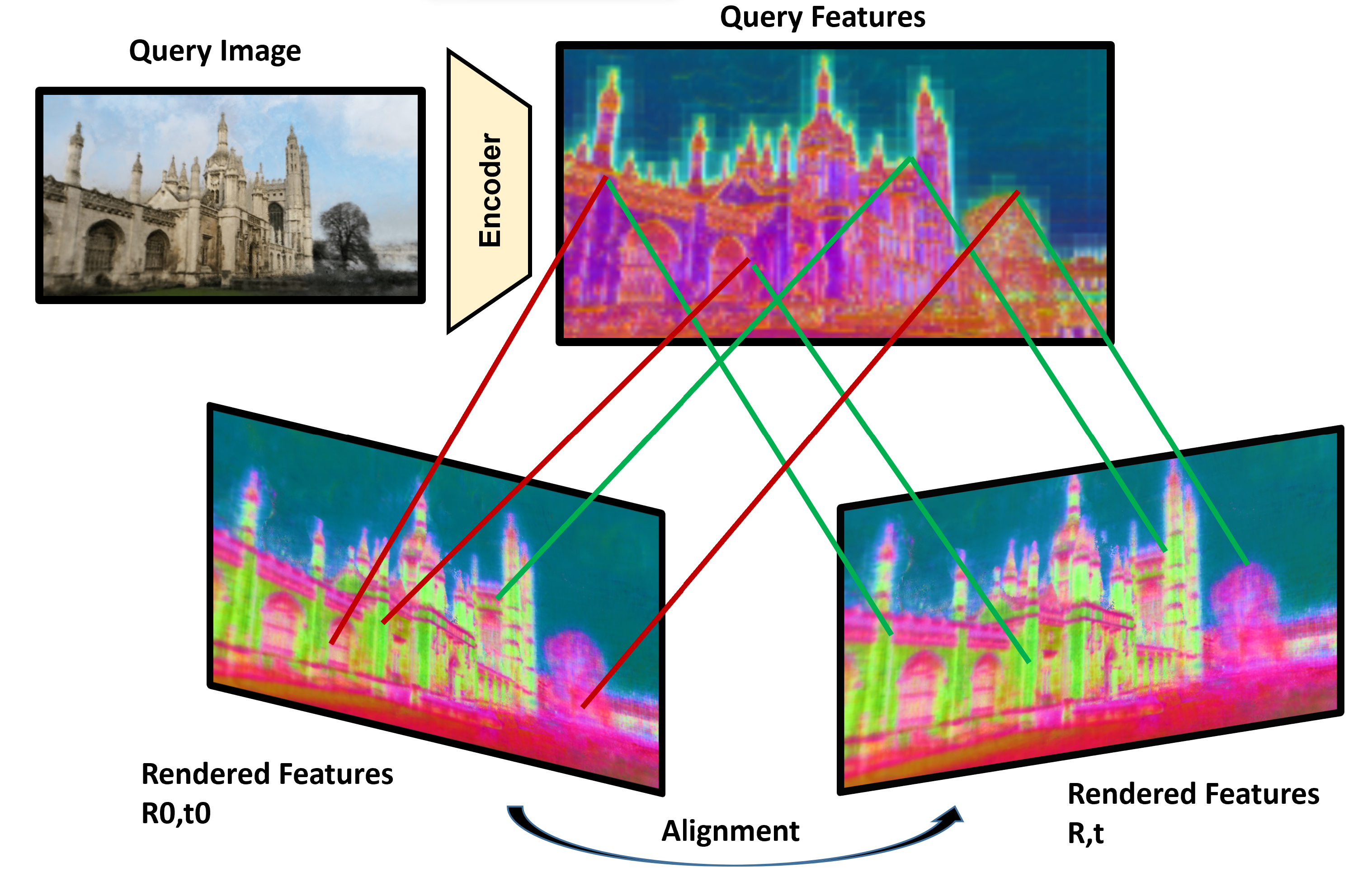}
\caption{Given an initial pose $R_0,t_0$, to estimate the pose of the query image, we align features from a query image to volumetric features which can be rendered at every position.}
\label{fig:teaser}
\end{figure}

Visual localization is the problem of estimating the precise position and orientation -- the camera pose -- from which a query image was taken in a known scene. 
Localization is a key part of advanced systems such as 
autonomous cars~\cite{HengICRA19ProjectAutoVisionLocalization3DAutonomousVehicle} and robots~\cite{LimIJRR15RealTimeMonocularImageBased6DoFLocalization}, and mixed-reality applications~\cite{ArthISMAR09WideAreaLocalizationMobilePhones,MiddelbergECCV14Scalable6DOFLocalization,LynenRSSC15GetOutVisualInertialLocalization}. 

Most existing visual localization approaches  differ on how they represent the scene, \eg, as a 3D Structure-from-Motion point cloud ~\cite{SeIROS02GlobalLocalizationDistinctiveVisualFeatures,IrscharaCVPR09FromSFMLocationRecognition,LiECCV10LocationRecPriorFeatureMatching,LiECCV12WorldwidePoseEst,SattlerICCV15HyperpointsFineVocabulariesLocRecogn,SattlerPAMI17EfficientPrioritizedMatching,SarlinCVPR19FromCoarsetoFineHierarchicalLocalization,TairaPAMI21InLocIndoorVisualLocalization,
Germain3DV19SparseToDenseHypercolumnMatchingVisLoc,HumenbergerIJCV22InvestigatingRoleImageRetrieval,SarlinCVPR21BackToTheFeature,von2020gn,vonstfumberg2020lmreloc,xu2020deep}, 
a database of images with known intrinsic and extrinsic parameters~\cite{ZhouICRA20ToLearnLocalizationFromEssentialMatrices,Zheng2015ICCV,Bhayani_2021_ICCV,Zhang06TDPVT,SattlerCVPR19UnderstandingLimitationsPoseRegression}, 
or implicitly as a set of weights of machine learning models such as random forests or neural networks
~\cite{ShottonCVPR13SceneCoordinateRegression,BrachmannICCV19ExpertSampleConsensusReLocalization,CavallariCVPR17OntheFlyCameraRelocalisation,KendallICCV15PoseNetCameraRelocalization,KendallCVPR17GeometricLossCameraPoseRegression,BrahmbhattCVPR18GeometryAwareLocalization,WangAAAI20AtLocAttentionGuidedCameraLocalization,Shuai3DV21DirectPoseNetwithPhotometricConsistency,ShuaiECCV22DFNetEnhanceAPRDirectFeatureMatching,MoreauCORL22LENSLocalizationEnhancedByNeRFSynthesis}.  
Yet, despite large algorithmical differences, these approaches have one thing in common: 
their scene representations are specifically built for localization and cannot be used for other tasks such as occlusion handling or path planning~\cite{Panek2022ECCV}.

Recent work shows that it is possible to use dense scene representations that can be used both for visual localization and other tasks~\cite{Panek2022ECCV,chen2023refinement,MoreauX23CROSSFIRECameraRelocImplicitRepresentation,GermainCWPRWS21FeatureQueryNetworks}.  
They require that the scene representation allows rendering the scene from given camera poses. 
These renderings can be either in the form of RGB-D images from which features for localization are extracted~\cite{Panek2022ECCV} or by directly rendering feature maps (stored in the scene representation) and scene depths~\cite{chen2023refinement,MoreauX23CROSSFIRECameraRelocImplicitRepresentation,GermainCWPRWS21FeatureQueryNetworks}. 
The latter type of methods uses an encoder that extracts the same features from the query image, thus facilitating feature matching between the query and the renderings. 
Prior work~\cite{chen2023refinement,GermainCWPRWS21FeatureQueryNetworks,tschernezki2022neural,kobayashi2022decomposing} thereby fixed the encoder and trained a neural implicit representation to render the same features using volumetric rendering~\cite{mildenhall2020nerf}. 

As shown in Fig.~\ref{fig:teaser}, this paper follows the approach of directly rendering features. 
In contrast to prior work, and similar to concurrent work~\cite{MoreauX23CROSSFIRECameraRelocImplicitRepresentation}, we show that better results can be obtained by jointly training the encoder with the features stored in the neural implicit representation instead of fixing the encoder (and thus the type of features) a-priori. 
In contrast to~\cite{MoreauX23CROSSFIRECameraRelocImplicitRepresentation}, we use the features for feature-metric pose refinement~\cite{vonstumberg2020lmreloc,SarlinCVPR21BackToTheFeature} rather than for explicit feature matching followed by pose estimation via RANSAC~\cite{Fischler81CACM} and a PnP solver~\cite{Haralick94IJCV,Kukelova13ICCV,Larsson2019ICCV}. 
In addition, we show that intermediate features from the part of the neural scene representation that predicts appearance and geometry provide a useful signal for learning the features used for localization. 
We show that our approach significantly outperforms~\cite{MoreauX23CROSSFIRECameraRelocImplicitRepresentation} on a commonly used outdoor dataset~\cite{KendallICCV15PoseNetCameraRelocalization}. 

In more detail, we propose to jointly learn the 3D layout of the scene and dense features for pose refinement. 
The features are learnt at a low overhead cost without extra supervision. 
Using recent developments in 3D surface reconstruction we learn an implicit field~\cite{mildenhall2020nerf} accurately capturing the geometry of the scene from RGB and depth supervision only. 
 {We propose to learn a joint embedding space, 
 for the image encoder and the features encoded in the  3D implicit field, and this in a self-supervised manner using contrastive learning.}
The field features are learned jointly together with the encoder, which removes the need for using a pre-trained feature extractor. 
Coupling the encoder features to the field features induces robustness to viewpoint changes as the field features are by definition robust to viewpoint changes (they only depend on the 3D position corresponding to the feature and not the direction from which this point is viewed). 
Furthermore, we exploit the fact that intermediate features of the implicit field encoding geometry and scene appearance carry some form of semantic meaning~\cite{Zhi2021iLabel}. 
More precisely, we leverage these features when learning the feature field for localization by linking them together. 
We show that this coupling of both fields improves localization performance. 
The resulting scene representation can be used to refine camera poses, both starting with inaccurate poses obtained via image retrieval and from more accurate poses. 
We show that our approach achieves state-of-the-art performance compared to other methods that represent the scene via a renderable neural implicit function~\cite{MoreauX23CROSSFIRECameraRelocImplicitRepresentation,chen2023refinement}. 

Our main contributions are:
    (1) In the context of using an implicit neural representation that encodes appearance, geometry, and features for localization, we show that the feature field and the corresponding image encoder can be learned jointly without supervision. 
    (2) We propose to couple the feature field with intermediate features of the branch predicting appearance and geometry. 
    We show the benefits of this approach experimentally. 
    (3) Integrating our approach into a visual localization pipeline based on feature-metric pose refinement, we show that our approach achieves state-of-the-art performance among methods using similar scene representations. 
    In particular, we outperform prior and concurrent work substantially on a standard outdoor benchmark. 
  
\section{Related Works}
 
\PAR{Neural implicit fields} A neural implicit field is a representation of a 3D {scene} whose underlying geometry and color are encoded through MLPs.  NeRF \cite{mildenhall2020nerf} use volumetric rendering~\cite{Kajiya1984VolumeRendering} to optimize the neural fields solely using RGB images with known intrinsics and camera poses.  
These methods perform very well novel view synthesis but are less accurate on surface reconstruction tasks as they do not directly optimize the 3D surface of the scene~\cite{yariv2021volume}. To that end, VolSDF \cite{yariv2021volume}, and NeuS \cite{wang2023neus} were the first to integrate signed distance fields (SDF) into the volume rendering equation of  \cite{mildenhall2020nerf}. 
In our work, we use NeuS~\cite{wang2023neus} 
 to make the induced feature field discriminative around the surface for accurate pose estimation.
Implicit models have been further used to learn and render feature fields. Using volumetric rendering, the fields are optimized by distilling 2D features through a feature loss. They use pre-trained 2D encoders such as DINO \cite{caron2021emerging} or CLIP \cite{radford2021learning} to infuse strong prior into the fields. These methods have shown great performances for scene manipulation and editing \cite{kobayashi2022decomposing,tschernezki2022neural}, semantic understanding \cite{kerr2023lerf}, and mapping in robotics scenarios \cite{shafiullah2022clipfields,mazur2022featurerealistic}.
Distilling 2D features into NeRFs has the attractive property of averaging information across multiple views. The resulting 3D features are thus shown to be more effective than their 2D counterparts. \cite{chen2023refinement} distills information from a pre-trained absolute pose regressor into a feature field. 
Several works also leverage this averaging process to distill noisy inconsistent 2D semantic observations into globally consistent 3D semantic field \cite{zhi2021inplace,fu2022panoptic,siddiqui2022panoptic}. Contrary to those methods, we do not use a pre-trained feature extractor, but rather train the extractor together with the feature field. 
Our feature field is jointly trained with the main field (representing 3D geometry and color information) and is optimized in a self supervised way, without the need for labelled data.

\PAR{Aligning images to scene representations}
Direct alignment {methods} optimize differences in pixel intensities by implicitly defining correspondences through the motion and geometry. 
Camera poses are then refined iteratively by minimizing the differences between an image and a projection of the scene into the current pose estimate. 
Such approaches are used in SLAM / Visual Odometry applications~\cite{alismail2017photometric,engel2014lsd,EngelPAMI17DirectSparseOdometry,schopscvpr19DADSLAM} or pose refinement \cite{SchopsCVPR17MultiViewStereo}. While being accurate, photometric alignment is inherently susceptible to variation in lightning condition or seasonal appearance changes. Furthermore, convergence often requires a good initialization. 
Learning deep features for feature-metric alignment can help to address the shortcomings of traditional photometric alignment, \eg, to increase robustness to illumination, viewpoint, and other types of changes. 
\cite{von2020gn,vonstumberg2020lmreloc,SarlinCVPR21BackToTheFeature,GermainCWPRWS21FeatureQueryNetworks,lindenberger2021pixelperfect,xu2020deep} thus optimize the alignment by minimizing the differences between features extracted from the image and projections of features stored in the scene. 
Their features are learned by casting the alignment as a metric learning problem. \cite{pietrantoni2023segloc} extends this framework by replacing deep features by learnt classification labels for more privacy preservation. 
Most related to this work, PixLoc~\cite{SarlinCVPR21BackToTheFeature} trains features for cross-seasonal wide-baseline alignment and learns deep features with a large basin of convergence through a hierarchical scheme. 
For the alignment, PixLoc relies on an explicit 3D model of the scene in the form of an SfM (Structure-from-Motion) point cloud. 
These points are used to project features extracted in reference images into the query images during refinement. 
Rather than explicitly transferring features from reference views into the query image, \cite{GermainCWPRWS21FeatureQueryNetworks} proposes to store features in a neural implicit field, termed a Feature Query Network (FQN). 
Still, \cite{GermainCWPRWS21FeatureQueryNetworks} requires an explicit 3D point cloud to query the field. 
The above-mentioned methods assume that the scene geometry is given and learn features based on the geometry. 
In contrast, our approach learns features jointly with a geometric representation of the scene and does not require an explicit model of the scene. 

\PAR{Visual localization} Given an initial pose estimate, using photometric or feature-metric camera pose optimization is one way to solve the visual localization problem~\cite{SarlinCVPR21BackToTheFeature}. 
A coarse initial estimate can be obtained via image retrieval~\cite{ToriiPAMI18247PlaceRecognitionViewSynthesis,ArandjelovicCVPR16NetVLADPlaceRecognition} against a database of reference images. 
The pose(s) of the top-retrieved image(s) then provide an approximation of the pose of the query image~\cite{ToriiICCVWS11VisualLocalizationByLinearCombination,HumenbergerIJCV22InvestigatingRoleImageRetrieval}. 
A more efficient alternative to image retrieval is to directly regress the camera pose using a neural network~\cite{KendallICCV15PoseNetCameraRelocalization,KendallCVPR17GeometricLossCameraPoseRegression,BrahmbhattCVPR18GeometryAwareLocalization,WangAAAI20AtLocAttentionGuidedCameraLocalization,Shuai3DV21DirectPoseNetwithPhotometricConsistency,ShuaiECCV22DFNetEnhanceAPRDirectFeatureMatching,MoreauCORL22LENSLocalizationEnhancedByNeRFSynthesis,ShavitECCV22LearningPoseAutoEncoders4ImprovingAPR,ShavitICCV21LearningMultiSceneAPRTransformers}. 
While initially not (significantly) more accurate than image retrieval~\cite{SattlerCVPR19UnderstandingLimitationsPoseRegression}, recent works have shown that pose accuracy can be significantly improved when using additional synthetic views, obtained via novel view synthesis, for training~\cite{Ng20223DV,MoreauCORL22LENSLocalizationEnhancedByNeRFSynthesis}. 
Yet, the most accurate poses are still obtained from 2D-3D correspondences between pixels in the query image and 3D points in the scene. 
These 2D-3D matches are used for pose estimation by applying a PnP solver~\cite{Haralick94IJCV,Kukelova13ICCV,Larsson2019ICCV} inside a RANSAC~\cite{Fischler81CACM} loop. 
These 2D-3D correspondences can be either computed explicitly via feature matching~\cite{SeIROS02GlobalLocalizationDistinctiveVisualFeatures,IrscharaCVPR09FromSFMLocationRecognition,LiECCV10LocationRecPriorFeatureMatching,LiECCV12WorldwidePoseEst,SattlerICCV15HyperpointsFineVocabulariesLocRecogn,SattlerPAMI17EfficientPrioritizedMatching,SarlinCVPR19FromCoarsetoFineHierarchicalLocalization,TairaPAMI21InLocIndoorVisualLocalization,
Germain3DV19SparseToDenseHypercolumnMatchingVisLoc} or by directly regressing them from the query image~\cite{ShottonCVPR13SceneCoordinateRegression,BrachmannICCV19ExpertSampleConsensusReLocalization,CavallariCVPR17OntheFlyCameraRelocalisation,brachmann2021visual,LiCVPR20HierarchicalSceneCoordinateClassification}. 
Our approach is based on feature-metric refinement of an initial pose. 
We show that we can improve both coarse pose estimates from image retrieval and more accurate poses obtained from 2D-3D matches. 

\PAR{Alignment and visual localization based on neural implicit fields} Our approach is based on a neural implicit field representation of the scene.
Leveraging the novel view synthesis capabilities of neural implicit fields, iNerf \cite{yenchen2021inerf} proposed to refine the pose of a reference image by photometrically aligning it with rendered images. 
In contrast to traditional alignment methods which optimize the objective using  the Gauss-Newton or Levenberg-Marquart algorithms, optimization is performed by directly backpropagating with gradient descent through the implicit neural field. 
\cite{lin2023parallel} extends this formulation by parallelizing the alignment process thanks to the faster INGP field model \cite{M_ller_2022}. 
This implicit field inversion formulation has been successfully integrated in SLAM frameworks \cite{sucar2021imap, zhu2022niceslam},  where photometric and depth alignment are used for camera tracking. 
Most relevant to our approach are the concurrent works from~\cite{MoreauX23CROSSFIRECameraRelocImplicitRepresentation,chen2023refinement}. 
\cite{chen2023refinement} performs feature-metric pose refinement by inverting a neural implicit field. 
Their method relies on a pre-trained encoder that extracts features from the query image. 
The field is trained to replicate the features provided by the encoder. 
In contrast, both our work and~\cite{MoreauX23CROSSFIRECameraRelocImplicitRepresentation} jointly learn the encoder and the feature field without extra supervision. 
In contrast to~\cite{MoreauX23CROSSFIRECameraRelocImplicitRepresentation}, we couple the feature field learned for localization with intermediate features of the neural field that encodes geometry and color information about the scene. 
While~\cite{chen2023refinement} and our approach perform feature-metric pose refinement, \cite{MoreauX23CROSSFIRECameraRelocImplicitRepresentation} uses the rendered feature maps for feature matching followed by PnP-based pose estimation. 
We show experimentally that our method outperforms both concurrent works, especially on outdoor scenes, thus validating our approach.

\begin{figure*}[tt]
\centering
\includegraphics[width=.9\linewidth]{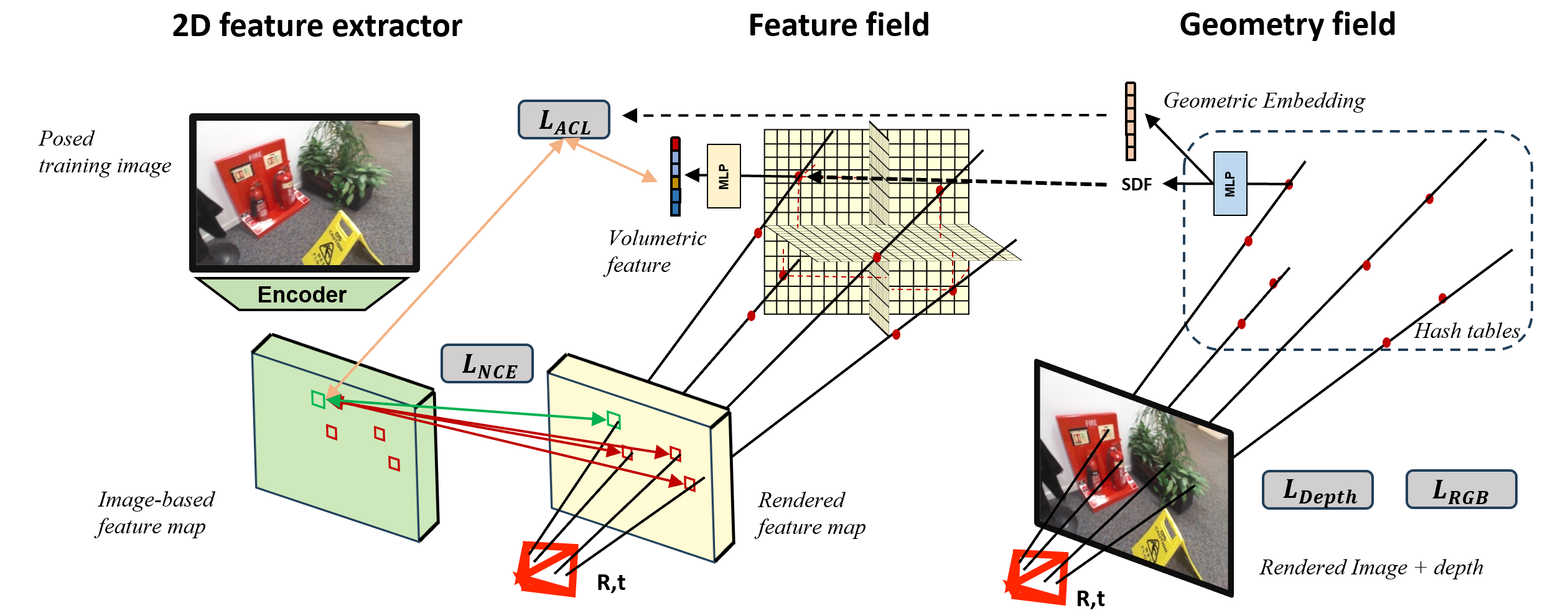}
\caption{ We jointly train an image encoder, a feature and a geometry field. The geometry field is trained through differentiable rendering with photometric and depth information. The feature field is trained by rendering features using the SDF from the geometry field and queried volumetric features. The rendered features are aligned with the image-based feature through a contrastive loss and a prototypical loss. }
\label{fig:pipeline}
\end{figure*}

\section{Method}
\label{sec:method}

In this section we describe how we jointly optimize a 3D feature field and a 2D dense encoder which share the same embedding space
o perform efficient pose refinement via feature metric alignment. 
In particular, Sec.~\ref{sec:geometry_field}  details how the underlying scene geometry is captured, Sec.~\ref{sec:feature_field} shows how the feature field is designed and optimized while in Sec.~\ref{sec:regul} we explain how we regularize and couple the feature field to the main field. Finally, in Sec.~\ref{sec:loc_pip}, the localization pipeline is presented.

\subsection{Background geometry field}
\label{sec:geometry_field}
 The geometry of the scene is encoded through an SDF function $f_{\theta}$. Each 3D point $x$ is mapped to a multi-resolution hash encoding $h = m_\phi(x)$~\cite{M_ller_2022} and decoded through the MLP to produce an SDF value $sdf$ and a geometric embedding $g$:
 $$sdf, g = f_\theta([x,\mu_p(x),h]),$$ where $\mu$ is a frequency positional encoding. The color is encoded through an MLP $f_\mu$ which takes as input the position $x$, the encoded viewpoint directions $\mu_d(d)$, the normals $n = \nabla_xs / \Vert \nabla_xs \Vert$ and the geometric embedding $g$ :
 $$c = f_\mu([x,\mu_d(d),n,b]).$$
 Following NeuS~\cite{wang2023neus}, the SDF function induces a density through the logistic density distribution $\phi_s = se^{-sx}/(1+e^{-sx})^2$ where $s$ is a learnable parameter. Volume rendering can thus be applied to train the SDF MLP $f_\theta$ to recover the surface. Given N samples from a ray parametrized by its origin and normalized direction $\{p_i = o + t_iv \,|\, i=1,...,n, t_i<t_{i+1} \}$, the color is rendered through $\widehat{C} = \sum_{i=1}^n T_i \alpha_i c_i$,  where $T_i$ is the accumulated transmittance 
 and $\alpha$  the discrete opacity value at sample $i$.

 Training the fields with RGB supervision only is fundamentally under-constrained, therefore we add depth supervision to better reconstruct textureless areas where photometric information would be too ambiguous. The rendered depth is also obtained through discrete integration of the weight distribution $\{w_i = T_i \alpha_i \}_i^n$ along the ray: $\widehat{D} = \sum_{i=1}^n T_i \alpha_i $. Having a properly captured underlying geometry better guarantees that the rendered feature maps are invariant with regard to any viewpoint change. Furthermore, the feature field will be mostly optimized around surface areas.    

\subsection{Feature field}
\label{sec:feature_field}

Along with the geometry and color, we jointly optimize a dense feature field and the feature extractor in a self-supervised manner,  without any additional supervision.
We adopt a triplane representation \cite{chen2022tensorf} for the feature field as it is an efficient 3D structure with good representational power and intuitive geometric interpretation. A triplane of resolution $P$ and feature dimension $d$ is composed of three orthogonal learnable feature planes $T_{xy},T_{yz},T_{xz}$ of dimension $P\times P\times n_{f}$ which intersect at the origin of the scene. A 3D point is mapped to a vector by projecting the point onto each plane and bilinearly interpolating a feature from each plane. The three resulting features are concatenated and decoded through a shallow MLP:  $u = f_\rho([u_{xy},u_{yz},u_{xz}])$ to further add a smoothness bias.
The feature field is not conditioned on the viewing direction as we want to obtain viewpoint invariant representations. Note that the underlying semantic information is viewpoint invariant.
Furthermore, as the feature extractor is not scale invariant we must account for the scale in the feature field. Hence, each pixel is modeled as a conical frustum. Following \cite{BarronICCV21ZipNeRF}, given a section $[t_{starts},t_{ends})$  along the ray, six points ${x_j}$ approximating the conical frustum shape in the section are sampled. Each point is associated to the mean of an isotropic Gaussian of standard deviation $\sigma_j = r_{t_j}/\sqrt{2}$, where $r_{t_j}$  
is the radius of the cone at ray distance $t_j$. To reduce aliasing, a down-weighting factor $w^j$ which approximates the fraction of the Gaussian distribution that is inside the cell of resolution $P$ is computed. As such, samples with a large scale compared to the resolution of the feature field will be down-weighted.
Each sample is projected onto the three planes, and the resulting 6*3 interpolated features are aggregated per plane before being fed to the MLP: 
\begin{align}
u(t_{starts},t_{ends}) = f_\rho([& mean(\{w^ju^j_{xy}\}_j),\nonumber \\ 
            &mean(\{w^ju^j_{yz}\}_j),\nonumber \\ 
            &mean(\{w^ju^j_{xz}\}_j)])
\end{align}
Given N samples from a ray, the rendered volumetric feature is obtained through alpha composition: 
 $$ V = \sum_{i=1}^N w_i u(t^i_{starts},t^i_{ends}) / \Vert u(t^i_{starts},t^i_{ends}) \Vert$$
The latter is further projected to the unit sphere. 

Through a contrastive framework, we jointly optimize the extractor and the feature field by aligning their features in a common embedding space. These representations benefit from the complementary inductive biases of the volumetric structure and the 2D encoder.
Given an input image, we first sample N pixels locations $\{u_i\}_{i=1}^n$, render their associated volumetric features $\{V_{u_i}\}_{i=1}^n$ from the feature field and interpolate their associated image-based features $\{E_{u_i}\}_{i=1}^n$ from the encoder. Through a InfoNCE loss \cite{oord2018representation}, we then maximize the similarity between rendered/image-based feature pairs associated with the same pixel while minimizing the similarity between rendered/image-based features associated with different pixels. Both volumetric/image similarities and image/volumetric similarities are optimized which leads to the following loss:

\begin{align}
L_{NCE} = -\frac{1}{2N} \sum_{i=1}^N &  \log\left(\frac{\exp{(E_{u_i} V_{u_i}}/\tau )}{ \sum_{j=1}^N \exp({E_{u_j} V_{u_i}/\tau})}\right)  
\nonumber \\ 
 & + \log\left( \frac{\exp{(V_{u_i} E_{u_i}/\tau}) }
 { \sum_{j=1}^N \exp({V_{u_j} E_{u_i}/\tau})} \right) 
\end{align}
where $\tau$ is a temperature parameter.  The co-visiblity between images and the stochasticity induced by the random sampling ensures a meaningful learning signal given the relatively low batch size.
 Pose refinement is an inherently local task which benefits from coarse to fine approaches and triplanes do not encode any frequency information. As such, to preserve both low and high-frequency encoding capabilities, we learn a coarse feature field with resolution $P_{coarse}$ and dimension $d_{coarse}$ as well as a fine feature field with resolution $P_{fine}$ and dimension $d_{fine}$. 
 We align the fine field against the first layer of the feature extractor which is sharper and the coarse field against the last layer of the feature pyramid module which produces smoother feature maps allowing the representations to capture scene information at different granularity level.

\subsection{Leveraging information from the main field}
\label{sec:regul}
To further regularize and constrain the learnt embedding space, we leverage information from the main geometric field in the form of correspondences and clustering assignments of geometric embeddings. 

\PAR{Two view consistency} The feature field has an inherent robustness to viewpoint change. To increase the robustness of the encoder to viewpoint changes we enforce multiview consistency between 2D extracted feature maps. 
Given a pair of images and N sampled pixels $\{u^1_i\} $ from the first image, the pixels are backprojected into 3D using the rendered depth at each pixel location. The reprojection $\{u^2_i\}$ of these 3D points in the second image is also backprojected into 3D. If the difference in 3D positions is below a threshold,  we consider that the 2D points form a correspondence, which allows for the handling of potential occlusions. 
We set the threshold value to 5cm  experimentally  as a compromise to allow for establishing correspondences even if the implicit model is not accurate enough while trying to minimize false positives. It is kept constant for all training scenes.  
We leverage the full batch of N samples and apply a ranking loss over the set of correspondences $C = \{u_j^1,u_j^2 \}_{j=1}^B$ and their image-based features interpolated at these pixel locations denoted by $\widehat{E}=\{E^1_{u_j^1},E^2_{u_j^2}\}_{j=1}^B$. 
Hence we minimize a mean Average Precision loss ($L_{mAP}$) over the set of corresponding image-based features: $L_{mAP} = \frac{1}{B}\sum_{j=1}^B \left(1 - AP(j,\widehat{E})\right)$.
The ranking loss enforces similarity between corresponding features over the whole batch which improves discriminativeness along with consistency on the encoder side. 

\PAR{Clustering geometric embeddings} 
Due to implicit field's ability to average information across views and propagate information \cite{zhi2021inplace}, the geometric embedding $g$ implicitly encodes rich color and geometric information. 
As such we use it to infuse priors into the embedding space with an auxiliary contrastive loss. By clustering the integrated geometric embeddings $G$, we can  store a set of K centroids $\{\widehat{G}_{k}\}_{k=1}^K$.
For a training iteration, given a batch of $N$ pixels associated with triplets of image-based feature, rendered feature and rendered geometric embedding $E_i,V,G_i\}_i$, we can assign each pixel to one of the K clusters, the assignment being given by  $c_i=\argmax_k sim(G_i,G_{c_k})$.
For each cluster $k$, mean image based feature $\widehat{E}_{k} = \sum_{i=1}^N E_i \mathbbm{1}_{c_i=k}$ and rendered based feature $\widehat{V}_{k} = \sum_{i=1}^N V_i \mathbbm{1}_{c_i=k}$ are also computed.  With these, we formulate our Auxiliary Contrastive Loss ($L_{ACL}$) as follows:
\begin{align}
L_{ACL}  =  -\frac{1}{2N} & \sum_{i=1}^N   \log\left(\frac{\exp{( \widehat{V}_{c_i} E_i}/\phi_{c_i} )}{ \sum_{j=1}^N \exp({ \widehat{V}_{c_j} E_i /\phi_{c_j}})}\right)  
\nonumber \\ 
 & + \log\left( \frac{\exp{( \widehat{E}_{c_i} V_i/\phi_{c_i}}) }
 { \sum_{j=1}^N \exp({ \widehat{E}_{c_j} V_i /\phi_{c_j}})} \right) 
\end{align}
where $\phi_{c_i}$ is the average distance to the cluster center $c_i$ introduced in \cite{li2020prototypical}.
Concretely, instead of aligning a feature to its counterpart from the other modality as in $L_{NCE}$, we align it to the associated mean vector from the other modality based on the geometric embeddings of cluster assignments. This forces the representations to focus on higher level information/structure conveyed by the prototypes.

\subsection{Localization pipeline}  
\label{sec:loc_pip}
Given a query image whose pose is unknown, we extract its image-based feature map $\widehat{F}$ with the encoder. From any pose $P_0 = [R_0\|t_0] \in SO(3)\times \R^3$ the volumetric-based feature map $\widehat{V}$ may be rendered by querying the feature field. As such, we want to find the pose that minimizes the feature metric distance between the rendered feature map and the image-based feature map. 
First, the most relevant database image is retrieved based on global descriptor similarity. Its pose acts as the initialization for the optimization. 
Our localization pipeline is agnostic to the type of global descriptor used for retrieval although refinement performance will depend on the accuracy of the initial retrieved pose (see 
Sec.\ref{sec:init} of supp. mat. for more details).
At each iteration, N pixels are uniformly sampled over the query image space $\{u_i,v_i\}_{i=1}^n$ and for each location the corresponding image 
feature $E_i$ is  bilinearly interpolated 
and the  volumetric feature  $V_i$ rendered. The average of their similarities is maximized with regard to the pose $$\min_{\mathbf{P}} \sum_{i=1}^N \left(1 - \frac{E_i^tV_i}{\Vert E_i \Vert \Vert V_i \Vert} \right)$$ by backpropagating with gradient descent from the rendered feature through the feature field back to update the input pose $\mathbf{P}$ on the $so(3)\times \R^3$ Lie algebra.
For the first half of the refinement iterations, we query the coarse feature field while, for the last second half of the refinement iterations, we query the fine field. This hierarchical approach increases the convergence basin of the pose refinement procedure as coarse feature field is smoother than the fine feature field.
The steps of the localization pipeline are summarized in the
Alg.\ref{alg:test} in the supp. mat.

\begin{table*}[tt!]
\scriptsize
\resizebox{\linewidth}{!}{
  \begin{tabular}{cl||c|c|c|c|c|c|c|c}
   &  Model  & Chess & Fire & Heads & Office & Pumpkin & Redkitchen & Stairs  & Average\\
    &  \multicolumn{8}{c}{Median pose error (cm.) ($\downarrow$), Median angle error (°) ($\downarrow$), Recall at 5cm/5° ($\%$) ($\uparrow)$} \\
    \hline
   \multirow{2}*{\begin{sideways} IR \end{sideways}}  
    & Oracle & 17.4/12.25 & 26.6/13.51 & 12.5/14.86 & 20.9/11.44 & 20.4/13.21 & 19.1/14.26 & 18.6/17.92 & 19.4/13.9 \\ 
   & DenseVLAD~\cite{ToriiPAMI18247PlaceRecognitionViewSynthesis} & 21.9/12.13 & 34.4/13.23 & 15.8/14.96 & 28.6/11.06 & 31.4/10.81 & 29.4/11.9 & 26.2/15.81 & 26.8/12.8 \\
   \hline
 \multirow{3}*{\begin{sideways} SOTA \end{sideways}}  
 & HLoc~\cite{SarlinCVPR19FromCoarsetoFineHierarchicalLocalization} & 0.8/0.11/100 & 0.9/0.24/99.4 &  0.6/0.25/100 & 1.2/0.20/100 & 1.4/0.15/100 & 1.1/0.14/98.6  &2.9/0.80/72.0 & 1.27/0.27/95.7 \\
   & DSAC*\cite{brachmann2021visual} & 0.5/0.17/99.9 &  0.8/0.28/98.9 & 0.5/0.34/99.8 & 1.2/0.34/98.1 & 1.2/0.28/99.0 & 0.7/0.21/97.0 & 2.7/0.78/92 & 1.09/0.34/97.8 \\
  & NBE+SLD~\cite{do2022learning} & 0.6/0.18/100 & 0.7/26/99.6 & 0.6/0.35/98.4 & 1.3/0.33/95.8 & 1.5/.33/94.4 & 0.8/0.19/96.6 & 2.6/0.72/85.2 &  1.16/4.01/95.7 \\

  \hline
    \multirow{3}*{\begin{sideways} APR$^{\textrm{IF}}$ \end{sideways}} 
   & MS-Transformer~\cite{ShavitICCV21LearningMultiSceneAPRTransformers} &  11/6.38 & 23/11.5 & 13/13.0 & 18/8.14 & 17/8.42 & 16/8.92 & 29/10.3 & 18.1/9.52\\
   & PAE~\cite{ShavitECCV22LearningPoseAutoEncoders4ImprovingAPR} & 13/6.61 & 24/12.0 & 14/13.0 & 19/8.58 & 17/7.28 & 18/8.89 & 30/10.3 & 19.3/9.52\\
   &  DFNet~\cite{ShuaiECCV22DFNetEnhanceAPRDirectFeatureMatching} &  3/1.12 & 6/2.30 & 4/2.29 & 6/1.54 & 7/1.92 &  7/1.74 & 12/2.63 & 6.4/1.93\\
    \hline
  \multirow{3}*{\begin{sideways} ILF  \end{sideways}}
   & NeFeS~\cite{chen2023refinement} (DFNet) & 2/0.79 & 2/0.78 & 2/1.36 & 2/0.60 & 2/0.63 & 2/0.62 & 5/1.31 & 2.43/0.87 \\
&     Ours (DV) & 1/0.22/92.90 & 0.8/0.28/90.93 & 0.8/0.49/71.20 & 1.7/0.41/81.20 & 1.5/0.34/85.65 & 2.1/0.41/75.32 & 6.5/0.63/49.0 & 
2.05/0.4/78.4 \\
 &    Ours (oracle)  & 0.8/0.2/94.39 & 0.9/0.30/98.30 & 0.7/0.42/76.00 & 1.5/0.38/82.45 &1.5/0.34/82.10 & 1.4/0.32/77.24 & 28.0/0.76/34.80 & 4.97/0.39/77.9\\
 & Ours (HLoc) & 0.5/0.10/100 & 0.5/0.21/99.40 & 0.4/0.25/100 & 0.7/0.19/99.62 & 0.6/0.15/100 & 0.6/0.14/98.86 & 2.5/0.67/76.20  & \textbf{0.83/0.24/96.3}\\
     \hline
    \end{tabular}
}
\caption{Refinement 7-Scenes, test set. Here, when we evaluate localization accuracy, we use the SFM based  pseudo GT references  from \cite{BrachmannICCV21OnTheLimitsPseudoGTVisReLoc}. For the pose refinement-based methods we indicate in parenthesis which method was used to initialize the pose. IR refers to image retrieval methods,  APR$^{\textrm{IF}}$ means Absolute Pose Regression improved by implicit fields such as NeRF,  ILF means using implicit features to align the query image with the implicit field.}
      \label{tab:7scenesWithpGT}
    \end{table*}

\section{Experiments}

In this section we provide further details regarding the training, implementation  details, and the datasets on which we evaluated our method
in Sec.~\ref{sec:impl}.
 Then, in  Sec.~\ref{sec:expres}, we provide an ablation study and discuss our experimental results. 

\subsection{Experimental Setup}
\label{sec:impl}

\PAR{Implementation details}
Training is performed end-to-end on a single NVIDIA-v100 32GB by minimizing the summation of the aformentioned losses.
We sample 1024 rays per training iteration and train over 100 thousand iterations. We optimize one field with its corresponding feature encoder per scene. A NeuS scheduler~\cite{wang2023neus} with 5000 steps of warmup is used. The ray sampling follows the HF-Neus formulation \cite{wang2022hfneus}. For a given ray, we use 64 uniform samples and 64 importance samples with a single up-sampling step. 
The full training process is described in Alg.\ref{sec:init} in supp. mat and further details may be found in Sec.~\ref{sec:impl_supp}.

\PAR{Datasets} We evaluate our localization approach on two real world localization datasets, 7-Scenes~\cite{ShottonCVPR13SceneCoordinateRegression} (with SfM and DSLAM pseudo ground truths)~\cite{BrachmannICCV21OnTheLimitsPseudoGTVisReLoc} and Cambridge Landmarks~\cite{KendallICCV15PoseNetCameraRelocalization}. 
These datasets are the ones mainly used by the learning-based visual relocalization methods, especially the ones using implicit map representations in their pipeline~\cite{GermainCWPRWS21FeatureQueryNetworks,Shuai3DV21DirectPoseNetwithPhotometricConsistency,ShuaiECCV22DFNetEnhanceAPRDirectFeatureMatching,MoreauCORL22LENSLocalizationEnhancedByNeRFSynthesis,MoreauX23CROSSFIRECameraRelocImplicitRepresentation}. 
We only use the depth to train the models, at inference time we only need the query RGB image.

\begin{table*}[tt!]
\scriptsize
\resizebox{\linewidth}{!}{
  \begin{tabular}{cl||c|c|c|c|c|c|c|c}
   &  Model  & Chess & Fire & Heads & Office & Pumpkin & Redkitchen & Stairs & Average \\
    & &  \multicolumn{8}{c}{Median pose error (cm.) ($\downarrow$), Median angle error (°) ($\downarrow$), Recall at 5cm/5° ($\%$) ($\uparrow)$} \\
    \hline
   \multirow{2}*{\begin{sideways} IR \end{sideways}}  
   & Oracle &  16/12.3 & 26/13.6 &  12/14.7 & 20/11.5 & 19/14.0 & 18/15.0 & 17/18.1  & 18.3/14.2 \\
 &DenseVLAD~\cite{ToriiPAMI18247PlaceRecognitionViewSynthesis}  &
 21/12.5 & 33/13.8 & 15/14.9 & 28/11.3  & 31/11.3 & 30/12.3 & 25/15.8  & 26.1/13.1 \\
    \hline
     \multirow{6}*{\begin{sideways} SOTA \end{sideways}}  
     & AS~\cite{SattlerPAMI17EfficientPrioritizedMatching} & 4/2 & 3/1.5 & 2/1.2 & 3/1.1 & 5/1.6 & 4/1.3 & 9/2.5 & 4.3/1.6 \\
   & HLoc~\cite{SarlinCVPR19FromCoarsetoFineHierarchicalLocalization} & 2.4/0.77/94.2 & 1.8/0.75/93.7 & 0.9/0.59/99.7 & 2.6/0.77/83.2 & 4.4/1.15/55.1 & 4.0/1.38/61.9  & 5.1/1.46/49.4  & 3/0.98/76.7\\
   & Kapture R2D2 (APGeM) \cite{humenberger2020robust} & 2/0.8 & 2/0.8 & 1/0.7 & 3/0.8 & 3/1.1 & 4/1.1 & 4/1.1 & 2.7/0.9 \\
  &  DSAC*~\cite{brachmann2021visual} &  1.8/0.59/97.8 & 1.7/0.77/94.5 & 1.0/0.66/98.8 & 2.7/0.79/83.9 & 3.9/1.05/62 & 3.9/1.24/65.5 & 3.5/0.93/78.0 & 2.6/0.86/82.9\\
 & NBE+SLD~\cite{do2022learning} &   2.2/0.75/93.7 & 1.8/0.73/94.1 & 0.9/0.68/96.6 & 3.2/0.91/74.8 & 5.6/1.55/44.6 & 5.3/1.52/45.7 & 5.5/1.41/44.6 & 3.5/1.07/70.6\\
   & HACNet~\cite{LiCVPR20HierarchicalSceneCoordinateClassification} &  2/0.7/97.5 & 2/0.9/96.7 & 1/0.9/100 &  3/0.8/86.5 & 4/1.0/59.9 & 4/1.2/65.5 & 3/0.8/87.5  & 2.7/0.9/84.8  \\
    & PixLoc~\cite{SarlinCVPR21BackToTheFeature}  (DV)  & 2/0.80 & 2/0.73 & 1/0.82 & 3/0.82 & 4/1.21 & 3/1.20 & 5/1.30 & 2.85/0.98 \\
\hline 
 \multirow{4}*{\begin{sideways} APR$^{\textrm{IF}}$ \end{sideways}} & Direct-PN~\cite{Shuai3DV21DirectPoseNetwithPhotometricConsistency} &  10/3.52 & 27/8.66 & 17/13.1 & 16/5.96 & 19/3.85 & 22/5.13 & 32/10.6 & 20.4/7.26 \\
&  MS-Transformer~\cite{ShavitICCV21LearningMultiSceneAPRTransformers} & 11/4.66 &  24/9.60 & 14/12.2 & 17/5.66 & 18/4.44 & 17/5.94 & 17/5.94 & 16.9/6.92\\
 &  DFNet~\cite{ShuaiECCV22DFNetEnhanceAPRDirectFeatureMatching} &  5/1.88 & 17/6.45 & 6/3.63 & 8/2.48 & 10/2.78 &  22/5.45 & 16/3.29 & 12/3.71 \\
 & LENS~\cite{MoreauCORL22LENSLocalizationEnhancedByNeRFSynthesis} & 3/1.3 & 10/3.7 & 7/5.8 & 7/1.9 & 8/2.2 & 9/2.2 & 14/3.6 & 8.3/2.96\\
   \hline
    \multirow{3}*{\begin{sideways} ILF \end{sideways}}
& FQN~\cite{GermainCWPRWS21FeatureQueryNetworks} (FQN-PnP)  
& 4/1.3 & 10/3 & 4/2.4 &10/3 & 9/2.4 & 16/4.4 & 140/34.7 & 27.6/7.3 \\
   & CROSSFIRE~\cite{MoreauX23CROSSFIRECameraRelocImplicitRepresentation} (DV) &  1/0.4 & 5/1.9 & 3/2.3 & 5/1.6 & 3/0.8 & 2/0.8 & 12/1.9 &   \textbf{4.43/1.38} \\
   
   &  Ours (DV)  & 3.4/0.97/74.75 & 2.5/0.94/83.48 & 2.7/1.59/62.10 & 4.1/1.20/60.68 &6.2/1.53/37.50  & 6.7/2.10/35.74 & 13.1/1.68/36.50 & 5.5/1.43/55.8\\
     \hline

    \end{tabular}
}
\caption{Refinement 7-Scenes, test set. Here, when we evaluate localization accuracy, we use the  Depth SLAM KinectFusion based  pseudo  GT references from~\cite{ShottonCVPR13SceneCoordinateRegression}.
For the pose refinement-based methods we indicate in parenthesis which method was used to initialize the pose. IR refers to image retrieval methods,  APR$^{\textrm{IF}}$ means Absolute Pose Regression improved by implicit field such as NeRF,  ILF means using implicit features to align the query image with the implicit field.}
      \label{tab:7scenesWithDSlam}
    \end{table*}

\subsection{Experimental Results}
\label{sec:expres}

In this section, we first ablate the main components of our model to validate design choices. We then evaluate the accuracy of our model against implicit feature localization methods, APR methods combined with NeRF or other implicit fields and  state-of-the-art visual localization methods. Finally we provide visualisations and discuss the limitations of our models. 

\begin{figure}[tt]
\centering
\includegraphics[width=1.\linewidth]{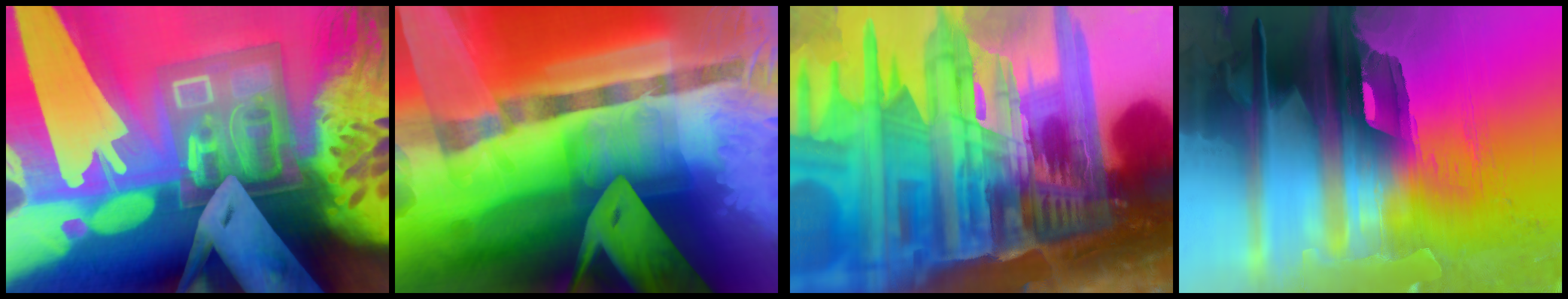}
\caption{Pairs of coarse feature map with $L_{ACL}$ (left) and coarse feature map without $L_{ACL}$ (right) for the same images.}
\label{fig:visu_prot}
\end{figure}

\PAR{Ablative study} In Tab.~\ref{tab:ablation},
we provide an ablative study of our model studying the influence of design choices while comparing single-scale refinement (coarse or fine) versus the hierarchical coarse-to-fine refinement. We note that integrating scale into the volumetric feature field facilitates the feature alignment with the non-scaled invariant 2D encoder (first \vs last row). 
Furthermore, using both $L_{mAP}$ and $L_{ACL}$ proves beneficial by enforcing consistency and structuring the embedding space versus using only one of the losses (second respectively third \vs last row). We further visualize the impact of $L_{ACL}$ in Fig.~\ref{fig:visu_prot}, which yields more structured feature maps while maintaining smoothness. 
The coarse level has a larger convergence basin than the fine level (larger recall), but often lacks the ability to refine up to fine accuracy (fourth \vs fifth row).
These levels are complementary for wide baseline pairs, hence the fine-to coarse refinement yields a significant gain over both.
Further ablations trying to bypass the feature field and remove depth are also available in 
Sec.\ref{sec:no_feature_field} of supp. mat. 

\begin{table}[h!]
    \centering

    \resizebox{\linewidth}{!}{
    \begin{tabular}{c|c|c|c|c||c|c}
     Scale & $L_{mAP}$ & $L_{ACL}$ & Coarse & Fine & Fire & KingsCollege  \\
     &  \multicolumn{5}{c}{Median pose error (cm.,°) ($\downarrow$)} \\
    \hline
    & \checkmark & \checkmark & \checkmark & \checkmark & 2.3/0.77/79.18 & 33.5/0.47 \\
    \checkmark & & \checkmark & \checkmark & \checkmark  & 1.7/0.53/83.51 & 35.5/0.79\\
    \checkmark & \checkmark & & \checkmark & \checkmark & 1.1/0.34/88.77 & 39.5/0.57\\
    \checkmark & \checkmark & \checkmark & \checkmark &  & 2.9/0.94/77.69 & 79.2/1.02 \\
    \checkmark & \checkmark & \checkmark & & \checkmark  & 1.2/0.37/82.84 & 164/1.25 \\
    \checkmark & \checkmark & \checkmark & \checkmark & \checkmark & 0.8/0.28/90.93 & 31.0/0.41\\
    \end{tabular}
    }
        \caption{Ablations of different model components on Fire (7-Scenes) and KingsCollege (Cambridge Landmarks). Models evaluated on KingsCollege in this table were trained with less iterations and less samples at each iteration than in Tab.~\ref{tab:cambridge}, which explains the discrepancy between the results.}
        \label{tab:ablation}
\end{table}

\begin{figure*}[t]
\centering
\includegraphics[width=1.\linewidth]{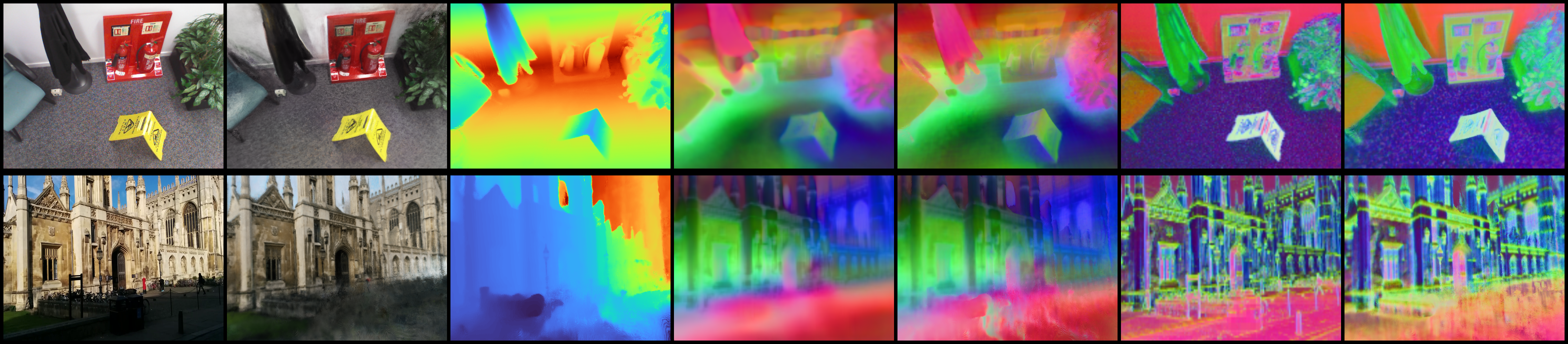}
\caption{From left to right, query image, rendered image, rendered depth, the PCA visualisations of encoded/rendered coarse feature maps, PCA visualisations of encoded/rendered fine feature maps. }
\vspace{-0.1cm}
\label{fig:visu}
\end{figure*}

\PAR{Comparisons with implicit feature representations-based methods}
We primarily compare our model against concurrent methods \cite{chen2023refinement,MoreauX23CROSSFIRECameraRelocImplicitRepresentation,GermainCWPRWS21FeatureQueryNetworks} based on neural implicit fields, which, similar to our method, synthesize features from an implicit field to perform localization. Furthermore, these methods also implicitly learn the 3D geometry of the scene during training. 
We report results for 7-Scenes SfM pseudo ground truth (pGT) in Tab.~\ref{tab:7scenesWithpGT}, 7-Scenes DSLAM pGT in Tab.~\ref{tab:7scenesWithDSlam}, and Cambridge Landmarks in Tab.~\ref{tab:cambridge}. 
On 7-Scenes SFM pGT, our method (Ours)
outperforms the implicit feature based alignment method NeFeS~\cite{chen2023refinement}
even when they start from the much more accurate APR poses predicted with DFNet~\cite{ShuaiECCV22DFNetEnhanceAPRDirectFeatureMatching} while we start from less accurate posse obtained by the Oracle (corresponding to the pose of the database image closest to the query image) or DenseVLAD~\cite{ToriiPAMI18247PlaceRecognitionViewSynthesis} (see the accuracy of these initial poses in the top part of  Tab.~\ref{tab:7scenesWithpGT}). This shows that our model
is able to better refine  in wide-baseline scenarios and has a larger convergence radius than NeFeS.

 Pose refinement is a local method and depends on the quality of the initialization, failure cases can often be avoided with a better initialization. This is illustrated on  one hand by the Oracle initialization (except on Stairs, on which the refinement failed for many images due to lack of texture and repetitive structure) \vs DenseVLAD initialization.  To further illustrate this, we initialize our refinement process with the highly accurate poses obtained with HLoc~\cite{SarlinCVPR19FromCoarsetoFineHierarchicalLocalization}
 and show
 that we can improve these initial poses to state-of-the-art
Further analyses on how  our model behaves when we
vary the initial poses  
can be found in 
Sec.\ref{sec:init} of supp. mat. 

On 7-Scenes, when we train our model with the DSLAM pGT, the noisiness of the poses makes it harder learn the underlying fields and therefore the method yields lower performances when refining (the same behavior was observed in \cite{chen2023refinement}). The performance of our method is comparable to the concurrent CROSSFIRE~\cite{MoreauX23CROSSFIRECameraRelocImplicitRepresentation} model: for some scenes our method performs better, on others and on average CROSSFIRE outperforms our model. 

Finally, on Cambridge Landmarks, our method obtains significantly better performances than other methods based on implicit feature fields, including CROSSFIRE. With NetVlad~\cite{ArandjelovicCVPR16NetVLADPlaceRecognition} pose initialization,  NeFeS obtains an average of (1.15m/1.3°) \vs  our (0.24m/0.48°) results.

\PAR{Comparison to other state-of-the art visual localization methods} 
Finally,  in each table  we also show state-of-the-art visual localization results, where the best solutions in general rely on 3D information used during training and/or inference and  the final pose is obtained either  with keypoint- or landmark matching-based methods~\cite{SattlerPAMI17EfficientPrioritizedMatching,SarlinCVPR19FromCoarsetoFineHierarchicalLocalization,SarlinCVPR21BackToTheFeature,humenberger2020robust,do2022learning} or  scene coordinates regression methods~\cite{brachmann2021visual} followed by  RANSAC~\cite{Fischler81CACM} and a PnP solver~\cite{Haralick94IJCV,Kukelova13ICCV,Larsson2019ICCV}. While these methods yield more accurate localization accuracy, our method similarly to other  
dense scene representations~\cite{Panek2022ECCV,chen2023refinement,MoreauX23CROSSFIRECameraRelocImplicitRepresentation,GermainCWPRWS21FeatureQueryNetworks}, has the further advantage 
that it can be used for other tasks, as they allow, for example, to render both images and depth from very different points of view (see \eg Fig.~\ref{fig:visu}).
We also report absolute pose regressor (APR) performances \cite{ShavitICCV21LearningMultiSceneAPRTransformers,ShavitECCV22LearningPoseAutoEncoders4ImprovingAPR,ShuaiECCV22DFNetEnhanceAPRDirectFeatureMatching} as they implicitly represent the scene from images with poses but as we can see, they yield much lower accuracy.

\PAR{Visualizations} In Fig.~\ref{fig:visu}, given a query image, we display renderings at the estimated query pose after refinement. Our model is jointly able to estimate accurate depth maps, faithful images, and feature maps for the refinement. The coarse feature map is smoother, allowing refinement from far initial estimates, while the fine feature map is sharper, allowing refinement to high accuracy. 3D-based rendered features and 2D-based encoded features are aligned, allowing the 2D-3D pose refinement to be effective. Volumetric and 2D feature maps are displayed side by side showing consistency between them. Uncertainty in the geometric field (last row, bottom right corner) leads to non aligned features which can prevent the refinement to converge to the ground truth pose.

\PAR{Limitations} The underlying field is only accurate within the volume defined by intersection of the frustums of the training camera poses. As such, query poses that are outside this volume and observe toward the scene boundaries may represent failure cases that our model has difficulty to refine, independently of the initialization. For the same reasons, drastically reducing the number of training images will decrease the quality of the underlying implicit model thus affecting localization accuracy.  
The difficulty of training the underlying implicit model on large real world scenes must be solved in order to scale up and localize on larger and more challenging datasets. For really large-scale maps, using a single geometric field would be insufficient, but we could envisage splitting the scenes as in~\cite{tancik2022blocknerf}. Depth (measured, rendered or estimated with monocular depth estimators) is also required, but  only for the training.
Finally, the iterative optimization procedure requires rendering features at every step which induces a large computational overhead (further discussions and experiments may be found in 
Sec.\ref{sec:impl_supp} in supp. mat.). This could be further optimized, but was considered out of the scope of this paper.

\begin{table}[t!]
\resizebox{\linewidth}{!}{
  \centering
  \begin{tabular}{cl|c|c|c|c|c}
   &  Model & King's & Old & Shop & St. Mary's & Average\\
    & & \multicolumn{4}{c|}{Median pose error (m.,°) ($\downarrow$)} \\
    \hline
\multirow{2}*{\begin{sideways} IR \end{sideways}}  
   & Oracle &   1.37/7.2 & 3.23/8.3 & 13.3/7.8 & 2.04/8.1 & 4.98/7.85 \\
    & NetVlad~\cite{ArandjelovicCVPR16NetVLADPlaceRecognition} & 2.9/5.9 & 4.05/7.5 & 1.36/7.20 & 2.87/9.36 & 2.8/7.49\\
& DenseVLAD~\cite{ToriiPAMI18247PlaceRecognitionViewSynthesis}  & 
 2.8/5.7 & 4.01/7.1 & 1.11/7.6 & 2.31/8  & 2.56/7.1 \\

    \hline
     \multirow{5}*{\begin{sideways}  SOTA \end{sideways}}
    & AS~\cite{SattlerPAMI17EfficientPrioritizedMatching} & 0.13/0.2 & 0.2/0.4 & 0.04/0.2 & 0.08/0.3 & 0.11/0.28 \\ 
    & HLoc~\cite{SarlinCVPR19FromCoarsetoFineHierarchicalLocalization} & 0.11/0.20 & 0.15/0.31 & 0.04/0.2 & 0.07/0.24 & 0.09/0.24 \\
    & Kapture R2D2 (APGeM) \cite{humenberger2020robust} & 0.05/0.1 & 0.09/0.2 & 0.02/0.1 & 0.03/0.1 & 0.05/0.1 \\
 &   DSAC*\cite{brachmann2021visual} & 0.15/0.3& 0.21/0.4 & 0.05/0.3 & 0.13/0.4 & 0.14/0.35 \\
 &    PixLoc~\cite{SarlinCVPR21BackToTheFeature}  (NV)   & 0.14/0.24 & 0.16/0.32 & 0.05/0.23 & 0.1/0.34  & 0.11/0.28 \\
& HACNet~\cite{LiCVPR20HierarchicalSceneCoordinateClassification} & 0.18/0.3 & 0.19/0.3 & 0.06/0.3 & 0.09/0.3 & 0.13/0.3 \\
 \hline
\multirow{3}*{\begin{sideways} APR$^{\textrm{IF}}$ \end{sideways}} 
&  MS-Transformer~\cite{ShavitICCV21LearningMultiSceneAPRTransformers} & 
0.83/1.47 & 1.81/2.39 & 0.86/3.07 & 1.62/3.99 & 1.28/2.73\\
 &  DFNet~\cite{ShuaiECCV22DFNetEnhanceAPRDirectFeatureMatching} 
 & 0.73/2.37 & 2/2.98 & 0.67/2.21 & 1.37/4.03 & 1.19/2.9
 \\
 & LENS~\cite{MoreauCORL22LENSLocalizationEnhancedByNeRFSynthesis} & 0.33/0.5& 0.44/0.9 & 0.27/1.6 & 0.53/1.6 & 0.39/1.2 \\
   \hline
     \multirow{6}*{\begin{sideways} ILF \end{sideways}}
 &  FQN~\cite{GermainCWPRWS21FeatureQueryNetworks} (FQN-PnP) 
 & 0.28/0.4 & 0.54/0.8 & 0.13/0.6 & 0.58/2.0 & 0.38/0.95 \\
 &   CROSSFIRE~\cite{MoreauX23CROSSFIRECameraRelocImplicitRepresentation} (DV)   & 0.47/0.7 & 0.43/0.7 & 0.20/1.2 & 0.39/1.4 & 0.37/1\\
   & NeFeS~\cite{chen2023refinement} (DFNet) & 0.37/0.62 & 0.55/0.9 & 0.14/0.47 & 0.32/0.99 & 0.35/0.75   \\
  &  Ours (NV) & 0.26/0.32 & 0.38/0.55 & 0.07/0.36 & 0.26/0.69 & \bf{0.24/0.48} \\
    &  Ours (DV) &  0.32/0.40 & 0.3/0.49 & 0.08/0.36 & 0.24/0.66 & \bf{0.24/0.48} \\
    \bottomrule
    \end{tabular}
}
\caption{Refinement Cambridge Landmarks.}
      \label{tab:cambridge}
    \vspace{-0.3cm}
    \end{table}

\section{Conclusion}
In this paper, we proposed a new representation for camera pose refinement that is learnt by aligning volumetric features and 2D features in a self-supervised way. These representations are jointly optimized alongside an implicit surface reconstruction model and benefit from the latter during training and inference. Indeed, pose refinement is a local problem which fundamentally benefits from the ability of spatially querying features at every location in the scene by leveraging an accurate underlying geometry field. Thus, jointly reasoning in 2D and 3D allows for better representation learning during training and hence facilitates localization during inference. Our approach does not require any SfM priors nor keypoint extraction and is grounded in 3D, which allows for effective visual localization. The model was evaluated on two real-world relocalization datasets.

\small
\PAR{Acknowledgments} Maxime and Torsten received funding from NAVER LABS Europe. Maxime was also supported by the SGS CVUT 2023 grant.

\clearpage
{
    \small
    \bibliographystyle{ieeenat_fullname}
    \bibliography{SL_Nif}
}

\appendix
\clearpage
\vspace{.5cm}
\noindent{\LARGE Appendix }
\vspace{.5cm}

In this supplementary material, we first analyze the influence of the accuracy of the initial pose estimate on the pose refinement process in Sec.~\ref{sec:init}. As further ablation, we try removing the feature field in Sec.~\ref{sec:no_feature_field} while in Sec.~\ref{sec:params} we evaluate the importance of the refinement parameters. Then, we show visualisations of the meshes in Sec.~\ref{sec:visu}. Finally, we provide further details about model architecture, training, and refinement parameters in Sec.~\ref{sec:impl_supp}.

\section{Initializing pose refinement}
\label{sec:init}
\begin{figure*}[t]
\centering
\includegraphics[width=.67\linewidth]{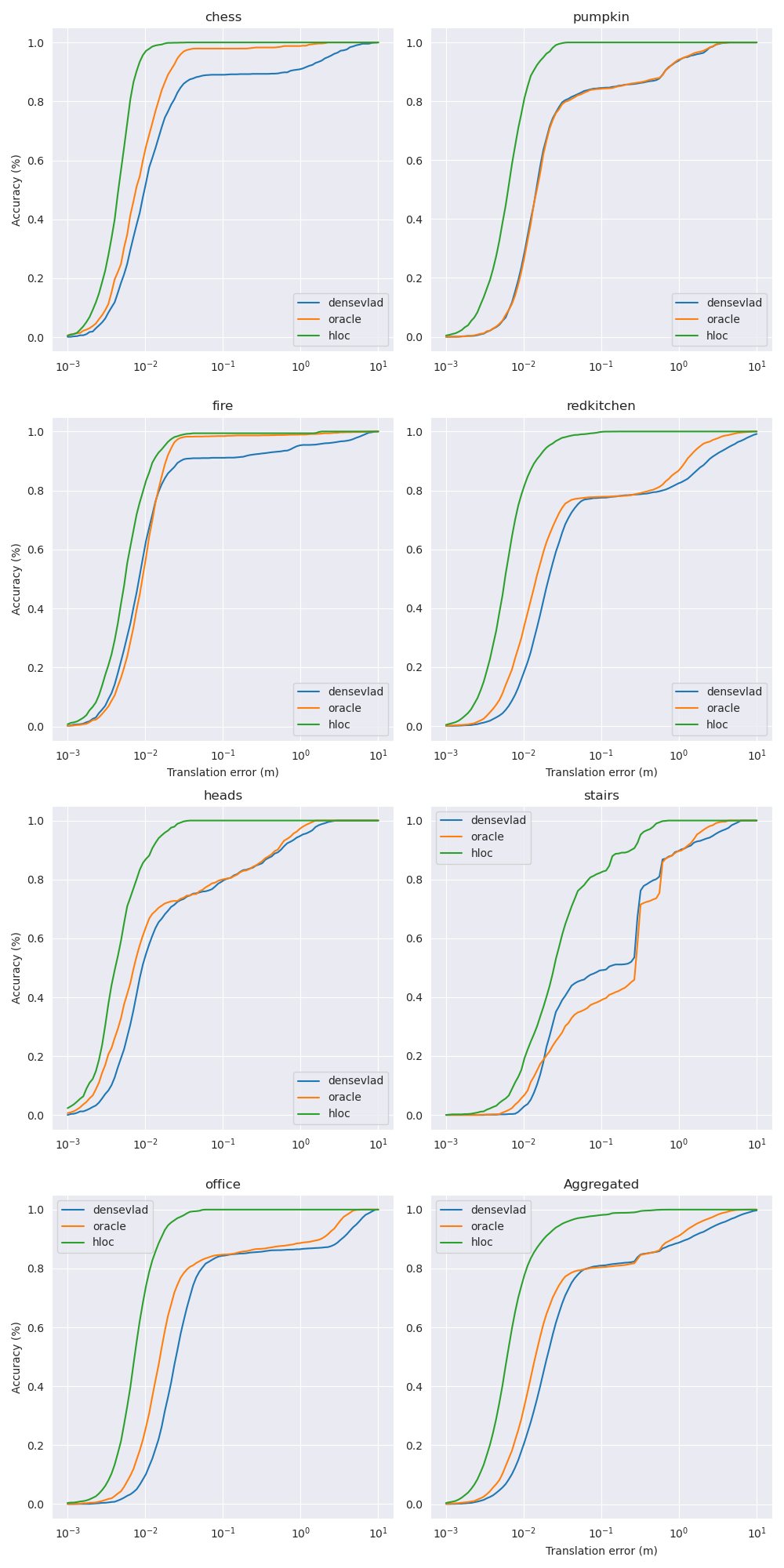}
\caption{Pose refinement on 7-Scenes ( SFM pGT) with multiple intializations: DenseVlad, Oracle, Hloc. For each translation error (x-axis) we plot the percentage of query images whose refined pose fall below the translation error (y-axis).}
\label{fig:recall}
\end{figure*}

The pose refinement procedure relies upon the quality of the initial pose estimate to succeed. If the latter falls into the basin of convergence of the optimization method, the pose will converge toward the correct query pose with a varying degree of accuracy. Otherwise the pose will diverge. In Fig.~\ref{fig:recall} we visualize cumulative recall plots for refinement with 3 different initializations (DenseVlad, Oracle, HLoc) for each scene from the the 7-Scenes dataset. Oracle initialization is the pose of the closest database image while DenseVlad initialization is the pose of the database image whose DenseVlad global descriptor has the most similarity with the query's DenseVlad global descriptor. Unsurprisingly, the more accurate the initial pose is, the higher the cumulative recall becomes. When initializing from DenseVlad/ Oracle, a plateau on the curve may be observed between the 10 cm and 1 meter meaning that no refined query pose lies between 10cm and 1 meter of its ground truth pose. Thus, the optimization either refines to fine accuracy or fully diverges whenever the query is looking outside the scene boundary or the initialization provide a bad estimate (which are the two critical failure cases). 
Refinement with HLoc initialization does not show this problem given the accuracy of the initial poses.

\begin{table}[b!]
    \centering
    \resizebox{\linewidth}{!}{
    \begin{tabular}{c|c|c|c|c||c|c}
     Feature Field & Gradient Flow & $L_{ACL}$ & Coarse & Fine & Fire & KingsCollege  \\
      \multicolumn{7}{c}{Median pose error (cm.,°) ($\downarrow$)} \\
    \hline
    &  &  & \checkmark &  &  7.5/2.66/29.50 & 831/7.94 \\
    &  & \checkmark & \checkmark &  & 7.3/2.48/32.61 & 507/5.32 \\
    & \checkmark &  & \checkmark &  & 3.4/1.20/54.47 & 208/3.21 \\
    & \checkmark & \checkmark & \checkmark &  & 2.8/0.98/55.60 & 175/3.47 \\
    \checkmark &  & \checkmark & \checkmark &  & 2.9/0.94/77.69 & 79.2/1.02 \\
    \hline
    \checkmark & & \checkmark & \checkmark & \checkmark & 0.8/0.28/90.93 & 31.0/0.41\\

    \end{tabular}
    }
        \caption{Feature field ablation on Fire (7-Scenes) and KingsCollege (Cambridge Landmarks).}
        \label{tab:ablation_field}
\end{table}

\begin{figure}[h!]                  
\centering                          
\includegraphics[width=1.\linewidth]{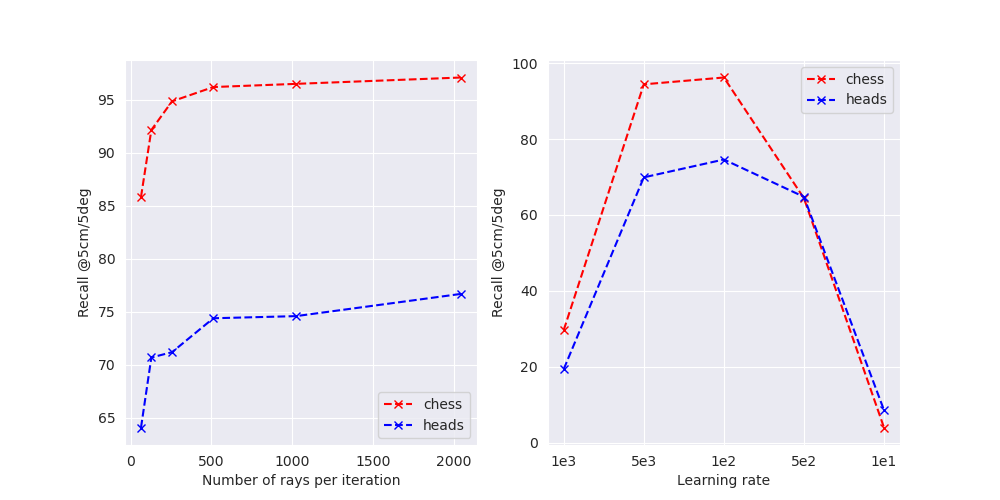}
\caption{Pose refinement on 7-scenes SFM pGT with varying number of rays sampled per iteration and initial learning rate.}
\label{fig:params}                    
\end{figure}

\section{Bypassing the feature field}
\label{sec:no_feature_field}
We provide further ablations by removing the feature field and replacing it with the geometric embedding of the geometric field. During training we align the geometric embedding integrated along rays and the 2D encoded features with $L_{NCE}$. In this setting, we further try ablating the prototypical loss $L_{ACL}$ and try letting the gradients flow ( Gradient Flow) to the geometric field by not detaching the gradients on the geometric embedding. Learning rates are lowered to facilitate training in this setup. We run the pose refinement by aligning the integrated geometric embedding to 2D encoded features on the coarse level only. Results are reported in Fig.~\ref{tab:ablation_field}. 
In both configurations the prototypical loss improves results (first \vs. second row, third \vs. fourth row), further reinforcing conclusions from Sec.\ref{sec:expres} of the main paper. 
Stopping the gradients from flowing during training hampers the alignment and degrades refinement accuracy ( second \vs. fourth row). Letting the gradient flow makes the representation learning less constrained and facilitates the alignment. However, the signal from $L_{NCE}$ also backpropagates to the hash encoding and geometric MLP, interferes with the reconstruction losses and degrades the quality of the underlying reconstruction. Indeed in this configuration, refinement accuracy is still lower than the feature field with solely the coarse level (fourth \vs. fifth row).
Overall it confirms that the representational power and inductive bias of a dedicated feature field is required to efficiently align volumetric features with 2D extracted features. 
We also ablate our model by training it without depth. As expected, the geometric field is less accurate in this case and the localization accuracy drops. For example in the case of Fire the median error increases from (0.8cm/0.28°) to (4.3cm/2.04°) or in the case of KingsCollege from (31cm/0.41°) to (35.7cm/0.48°). Note that no parameter tuning or adaptation as been done before removing depth, as such the performance drop could be much dampened but this is outside the scope of this paper. 

\section{Refinement parameters}         
\label{sec:params}            
For the pose refinement, the parameters of the optimizer can be adjusted ( learning rate and decay rate) as well as the number of ray sampled per iteration and the number of points per ray. We keep the same number of sampled points per ray as during training ( 128 samples), use a decay rate of 0.33 and study the remaining two parameters. We ran the pose refinement procedure on the Heads and Chess scenes from the 7-Scenes dataset, using the SFM pseudo ground truth ( SFM pGT) from ~\cite{BrachmannICCV21OnTheLimitsPseudoGTVisReLoc}, with different number of sampled rays per iteration and different initial learning rates for each query during the refinement. We report results in Fig.~\ref{fig:params}. Increasing the number of sampled rays provides more information during optimization and improves the resulting accuracy. Below a threshold number of 512 rays the optimization is hampered by the lack of coherent signal, while above 1024 rays accuracy gains become more marginal. On the other hand, increasing the number of sampled rays induces a significant computational overhead. We set the number of refinement steps to 300 ( 150 for the coarse level and 150 for the fine level), most queries require less than 100 steps to converge but this is necessary to refine more challenging queries. Overall the number of rays per iterations and the number of refinement iterations determine the localization time per query ( around 15s per query with 512 rays and 300 iterations on NVIDIA V100 32Gb GPU). As mentioned earlier, this could be improved, however optimizing the code is outside of this paper's scope.
Finally, the refinement is very sensitive to the initial learning rate of the optimizer which depends upon the optimization profile. The latter must be set carefully. 
                                    
\section{Visualisations}
\label{sec:visu}

\begin{figure*}[tt!]
\centering
\includegraphics[width=1.\linewidth]{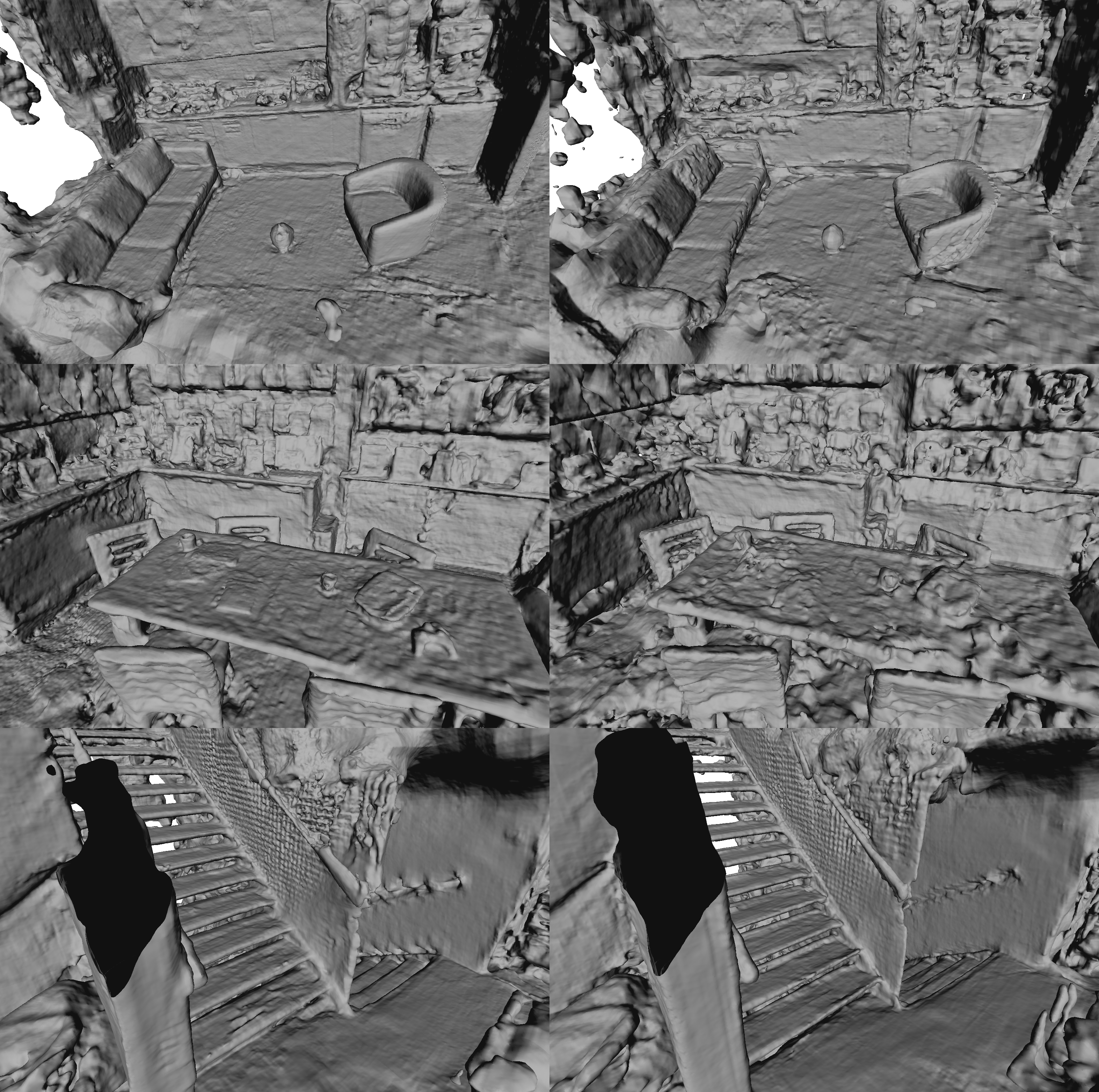}
\caption{Meshes of the scenes from the 7-Scenes dataset, obtained using the SFM pGT (left column) and the DSLAM pGT (right column). From top to bottom: Pumpkin, Red Kitchen, Stairs.}
\label{fig:meshes57}
\end{figure*}

\begin{figure*}[tt!]
\centering
\includegraphics[width=1.\linewidth]{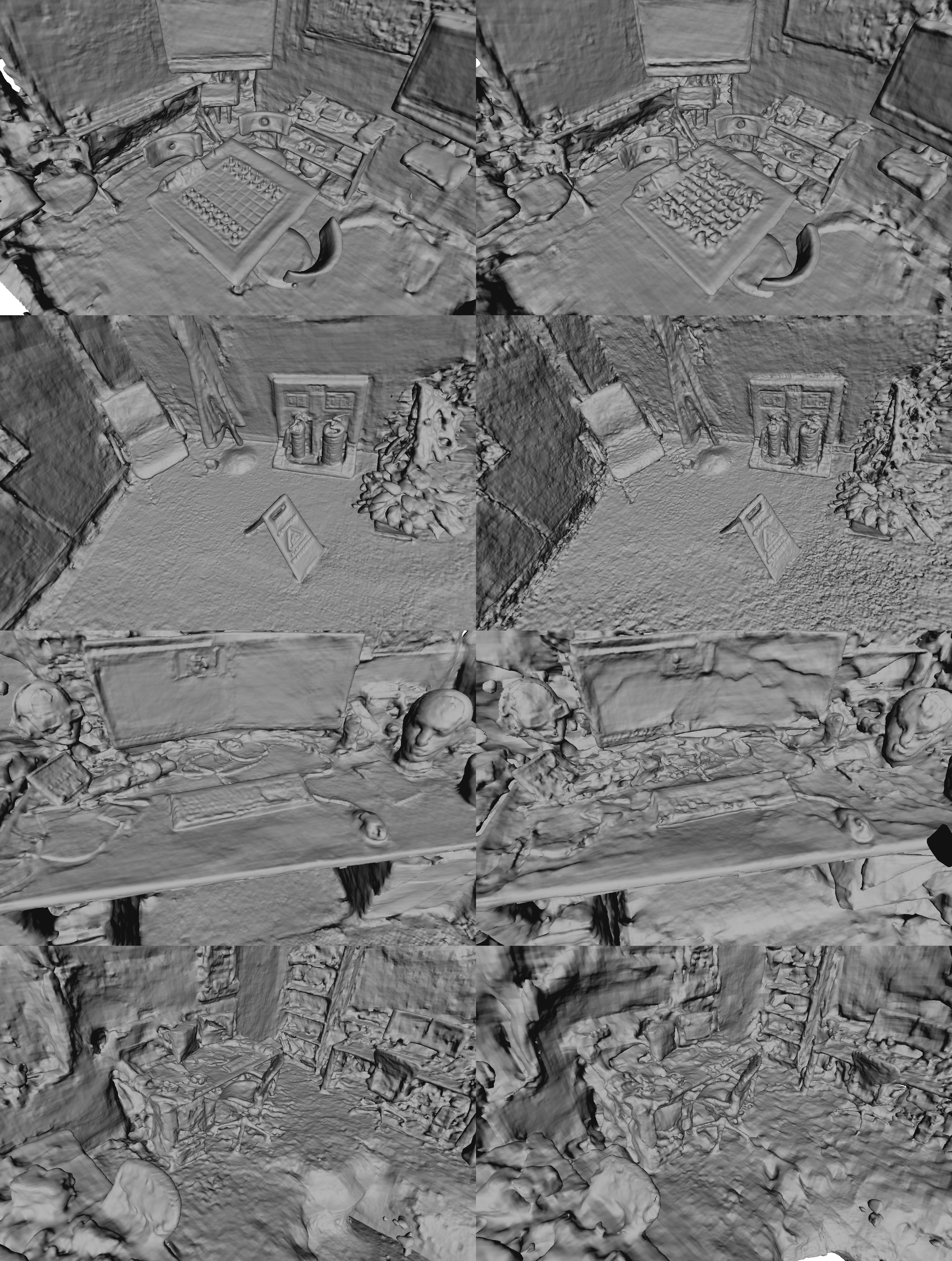}
\caption{Meshes of the scenes from the 7-Scenes dataset, obtained using the SFM pGT (left column) and the DSLAM pGT (right column). From top to bottom: Pumpkin, Red Kitchen, Stairs.}
\label{fig:meshes14}
\end{figure*}

In Fig.~\ref{fig:meshes14} and \ref{fig:meshes57} we show 3D meshes extracted with the marching cubes algorithm \cite{lorensen1998marching} from models trained on the SFM pseudo ground truth ( on the left column) and the DSLAM pseudo ground truth ( on the right column) considering the same training parameters settings. Visually, the SFM pGT meshes appear to capture much more surface details than the DSLAM pGT meshes while being more accurate. This corroborates the claim that the DSLAM pGT leads to inferior geometric models which directly correlates to a lower accuracy relative to other methods compared to using SFM pGT to train. It also confirms the necessity of having accurate poses to train implicit models. A line of work \cite{lin2021barf,bian2023nope,chen2023dbarf} tackles this issue by jointly optimizing the image poses and the field during training. However such techniques could not be applied to visual relocalization as it only optimizes the training poses. Hence, training and testing poses would be shifted and a new set of pseudo ground truth would need to be computed.

\section{Additional implementation details}
\label{sec:impl_supp}

\PAR{Geometric Field}
The geometric MLP has 8 layers of hidden dimension 256 with weight normalization, a skip connection at layer four, and softplus activations \cite{zheng2015improving}. The color MLP has 3 layers of hidden dimension 256 and RELU activations. The geometric field is trained with a learning rate of 3e-5 with an Adam optimizer. The temperature $\tau$ in the $L_{NCE}$ was kept to 1. Concerning the SDF, we follow \cite{atzmon2020sal} and initialize $f_\theta$ such that it produces an approximated SDF to the unit sphere, the sign is reversed depending if cameras are looking "inward" ( outdoor scenes: Cambridge Landmarks) or "outward" ( indoor scenes: 7scenes).

\PAR{Feature field}
The coarse field Triplane \cite{chen2022tensorf} has a resolution of 512 and feature dimension of 32, the coarse feature maps have a resolution of 128. The fine field Triplane has a resolution of 1024 and feature dimension of 32, the fine feature maps have a resolution of 96. 
The resolution of the feature field also correlates with the maximum accuracy that the pose refinement process can reach. While increasing the resolution for larger scale scenes helps, it comes at a computational and storage cost increase.
The feature field MLPs have 2 layers with a hidden dimension of 128.  The feature field is trained with a learning rate of 1e-4 also with an Adam optimizer.

\PAR{Encoder}
  The image-based feature extractor contains a transformer based encoder which outputs a set of feature maps at different scale and a convolution-based decoder that progressively fuses and upsamples them to obtain the final feature map at image input resolution. In particular, the encoder is based upon the SWIN-t transformer architecture \cite{liu2021swin} combined with a convolutional feature pyramid type of network \cite{lin2017feature}. The encoder is trained with a learning rate of 1e-4 with Adam optimizers \cite{kingma2014adam}.

\PAR{Pose refinement}
For the pose refinement, 512 rays are sampled per iteration on 7-Scenes and 1024 rays are sampled per iteration on Cambridge Landmark for a total of 300 refinement iterations per image. Optimization is ran through an Adam optimizer with initial learning rate of 1e-2 for 7scenes and 2e-2 for Cambridge Landmarks. The decay rate is set to 0.33 in both cases. We detail the steps of the localization pipeline in Alg.~\ref{alg:test}.

\PAR{Misc}
We minimize the loss $L = \lambda_{rgb}L_{rgb} + \lambda_{depth}L_{depth} + \lambda_{eik}L_{eik} + \lambda_{NCE}L_{NCE} + \lambda_{ACL}L_{ACL} + \lambda_{mAP}L_{mAP}$ with $\lambda_{rgb}=\lambda_{depth}=\lambda_{NCE}=\lambda_{mAP}=0.1$, $\lambda_{ACL}=0.05$ and $\lambda_{eik}$ annealed from 0.001 to 0.01. The $L_{eik}$ is a regularization loss enforcing that the gradients of the SDF function with regard to the input position have a norm of one. It constrains the SDF function to behave as a geometric distance~\cite{AmosICML2020ImplicitGeomReg}.
We further lay out the steps of the training pipeline in Alg.~\ref{alg:train}.  For simplicity, distinction between fine and coarse levels are omitted. In practice the losses are similar, only the resolution of the feature fields and the locations at which the image-based features are extracted from the encoder differ between the coarse and fine levels.
Optimizing the feature field should not affect the distribution of weights and the underlying geometry. Indeed letting the gradients flow would allow the networks to predict inconsistent features by dampering the geometry resulting in artifacts and inconsistent values. Therefore we detach gradients on the weights $w$ between the main field and the feature field so that no gradients propagates between fields. Theoretically, training could have been performed in two stages, the geometry field followed by the joint training of the encoder and the feature field. However to fully leverage the coupling and simplify the training procedure we chose to train everything at once. The total storage requirement for encoder and fields under the aforementioned settings is 2GB. 

 \begin{algorithm*}
 \caption{Training process.}
  \label{alg:train}
      
 \SetAlgoLined
 \KwData{Posed RGB images with associated depth}

 Initialize fixed size queue feature Q \\
 \For{iteration in range(N\_iterations)}{
    Sample a random pair of image $I^a,I^b$ \\
    Extract 2D image-based features $E_a,E_b$ from images \\
    Sample 1024 rays $\{r_n\}_{n=1}^{1024}$ in $I^a$, sample 128 points $\{p_i\}_{i=1}^{128}$ per ray \\
    Bilinearly interpolate $E_a,E_b$ at ray origin pixels which yields $\{E_a^n,E_b^n\}_{n=1}^{1024}$ \\
    Query geometry and feature fields to obtain color, geometric embedding, SDF and feature per 3D point $c_i^n,sdf_i^n,g_i^n,u_i^n$ \\
    Integrate along rays to obtain rendered colors, geometric embeddings, depths and features $C^n,G^n,D^n,V^n$ \\
    
    \If{iteration \% 50==0}{
        Compute sets of prototypes from Q
    }
    \Else{
        Add $\hat{G}^n$ to Q
    }
    
    Compute $L_{rgb},L_{depth},L_{eik}$ \\
    Compute $L_{NCE}$ from $E_a^n$ and $V^n$ (Sec.~\ref{sec:feature_field} main paper) \\
    Compute $L_{ACL}$ from $E_a^n,V^n$ and prototypes $\hat{G_n}^n$   (Sec.~\ref{sec:regul} main paper)\\
    Find 2D-2D tentative correspondences for rays $\{r_n\}_{n=1}^{1024}$, compute $L_{mAP}$ from $E_a,E_b$ interpolated at correspondence positions (Sec.~\ref{sec:regul} main paper)\\
    
    Jointly update fields and encoder parameters by minimizing  the overall loss $L = \lambda_{rgb}L_{rgb} + \lambda_{depth}L_{depth} + \lambda_{eik}L_{eik} + \lambda_{NCE}L_{NCE} + \lambda_{ACL}L_{ACL} + \lambda_{mAP}L_{mAP}$ \\
    }

\vspace{.5cm}    

\end{algorithm*}

 \begin{algorithm*}
 \caption{Localization pipeline.}
  \label{alg:test}
      
 \SetAlgoLined
 \KwData{Query image, trained fields and encoder}
 \KwResult{Estimated pose P of the query image}

 Retrieve initial pose P ( global descriptor or SFM localization)\\
 Extract 2D image-based query coarse and fine features $E_{qcoarse}, E_{qfine}$ from query image \\
 \For{granularity in [coarse, fine]}{
    \For{iteration in range(N\_ref\_iter)//2}{
    From pose P, sample 512 rays $\{r_n\}_{n=1}^{512}$ in $I^a$, sample 128 points $\{p_i\}_{i=1}^{128}$ per ray \\
    Bilinearly interpolate $E_{qgranulairty}$ at ray origins \\
    Query geometry and feature fields to obtain SDF and granularity feature per 3D point $sdf_i^n,u_i^n$ \\
    Integrate along rays to obtain rendered features $V^n$  \\
    Update P by maximizing cosine similarity between $V^n$ and $V_{qgranulairty}^n$ and backpropagating through the feature field (Sec.\ref{sec:loc_pip} main paper) \\
    }
 
 }

\vspace{.5cm}    
\end{algorithm*}

\end{document}